%% file: main.tex
\documentclass[10pt,twocolumn,letterpaper]{article}

\usepackage{iccv}
\usepackage{times}
\usepackage{epsfig}
\usepackage{graphicx}
\usepackage{amsmath}
\usepackage{amssymb}

\usepackage{pifont}
\usepackage{booktabs}
\usepackage{multirow}
\usepackage{capt-of}
\usepackage{makecell}
\usepackage{colortbl}
\usepackage{dsfont}
\usepackage{comment}
\usepackage[dvipsnames]{xcolor}

\usepackage[accsupp]{axessibility}  %

\usepackage{tocloft}

\makeatletter
\renewcommand{\numberline}[1]{%
  \@cftbsnum #1\@cftasnum\hspace*{1em}\@cftasnumb%
}
\makeatother

\usepackage[pagebackref=true,breaklinks=true,colorlinks,bookmarks=false]{hyperref}

\usepackage[capitalize]{cleveref}
\crefname{section}{Sec.}{Secs.}
\Crefname{section}{Section}{Sections}
\Crefname{table}{Table}{Tables}
\crefname{table}{Tab.}{Tabs.}

\renewcommand{\paragraph}[1]{\vspace{1.25mm}\noindent\textbf{#1}}

\definecolor{baselinecolor}{gray}{.9}
\definecolor{darkgreen}{rgb}{0.13, 0.55, 0.13}

\let\originalleft\left
\let\originalright\right
\renewcommand{\left}{\mathopen{}\mathclose\bgroup\originalleft}
\renewcommand{\right}{\aftergroup\egroup\originalright}

\iccvfinalcopy %

\def\iccvPaperID{5097} %

\begin{document}

\title{Neighborhood Commonality-aware Evolution Network for Continuous Generalized Category Discovery}

\author{%
  Ye Wang\textsuperscript{$1$} \qquad
  Yaxiong Wang\textsuperscript{$2$} \qquad
  Guoshuai Zhao\textsuperscript{$1$} \qquad
  Xueming Qian\textsuperscript{$1$} \\
  \textsuperscript{$1$}Xi'an Jiaotong University \quad \\
  \textsuperscript{$2$}Hefei University of Technology \quad \\
  \tt\small \{xjtu2wangye@stu,wangyx15@stu,guoshuai.zhao@, qianxm@mail\}.xjtu.edu.cn \\
}



\maketitle


\input{sections/0.abstract.tex}

\input{sections/1.introduction.tex}
\input{sections/2.related_work.tex}
\input{sections/3.problem_formulation.tex}

\input{sections/4.method.tex}
\input{sections/5.experiment.tex}

\input{sections/6.conclusion.tex}

\section*{Acknowledgements}
This work was supported by the NSFC under Grant 62272380 and 62103317.

{\small
\bibliographystyle{ieee_fullname}
\bibliography{ref}
}
\clearpage

\appendix




\end{document}

%% file: sections/0.abstract.tex
\begin{abstract}
Continuous Generalized Category Discovery (C-GCD) aims to continually discover novel classes from unlabelled image sets while maintaining performance on old classes. In this paper, we propose a novel learning framework, dubbed Neighborhood Commonality-aware Evolution Network (NCENet) that conquers this task from the perspective of representation learning. Concretely, to learn discriminative representations for novel classes, a Neighborhood Commonality-aware Representation Learning (NCRL) is designed, which exploits local commonalities derived neighborhoods to guide the learning of representational differences between instances of different classes. To maintain the representation ability for old classes, a Bi-level Contrastive Knowledge Distillation (BCKD) module is designed, which leverages contrastive learning to perceive the learning and learned knowledge and conducts knowledge distillation. Extensive experiments conducted on CIFAR10, CIFAR100, and Tiny-ImageNet demonstrate the superior performance of NCENet compared to the previous state-of-the-art method. Particularly, in the last incremental learning session on CIFAR100, the clustering accuracy of NCENet outperforms the second-best method by a margin of 3.09\% on old classes and by a margin of 6.32\% on new classes. Our code will be publicly available at \href{https://github.com/xjtuYW/NCENet.git}{https://github.com/xjtuYW/NCENet.git}.
\end{abstract}

%% file: sections/1.introduction.tex
\section{Introduction}
\label{sec:intro}

Category Discovery (CD)~\cite{vaze2022generalized,han2021autonovel} aims to discover novel classes in unlabelled images partially based on the knowledge learned from labelled images. This task has numerous applications in real-world scenarios, such as novel disease detection in medical images, new species discovery, and automatic image data annotation, etc. This paper focuses on a specific setting of Continuous Generalized Category Discovery (C-GCD)~\cite{wu2023metagcd}, \ie, given a sequence of unlabelled image sets, we need to continually discover novel categories from each unlabelled image set while maintaining performance on old categories. This task is quite challenging from many perspectives, such as the old training set being inaccessible during incremental learning sessions, the incremental image set being unlabeled, and the number of categories being unknown.

Directly applying the conventional CD methods can not solve this task well due to the following two reasons: 

1) Labelled data reliance. In existing CD methods, labelled data are often required to guide the learning of discovering novel classes in unlabelled data. 

2) Catastrophic forgetting issue. C-GCD is an incremental task that consists of multiple incremental learning sessions. As the learning process proceeds, the absence of old data in incremental learning sessions will drop the clustering performance of some CD methods significantly.

\begin{figure}[t]
    \centering
    \includegraphics[width=.9\columnwidth]{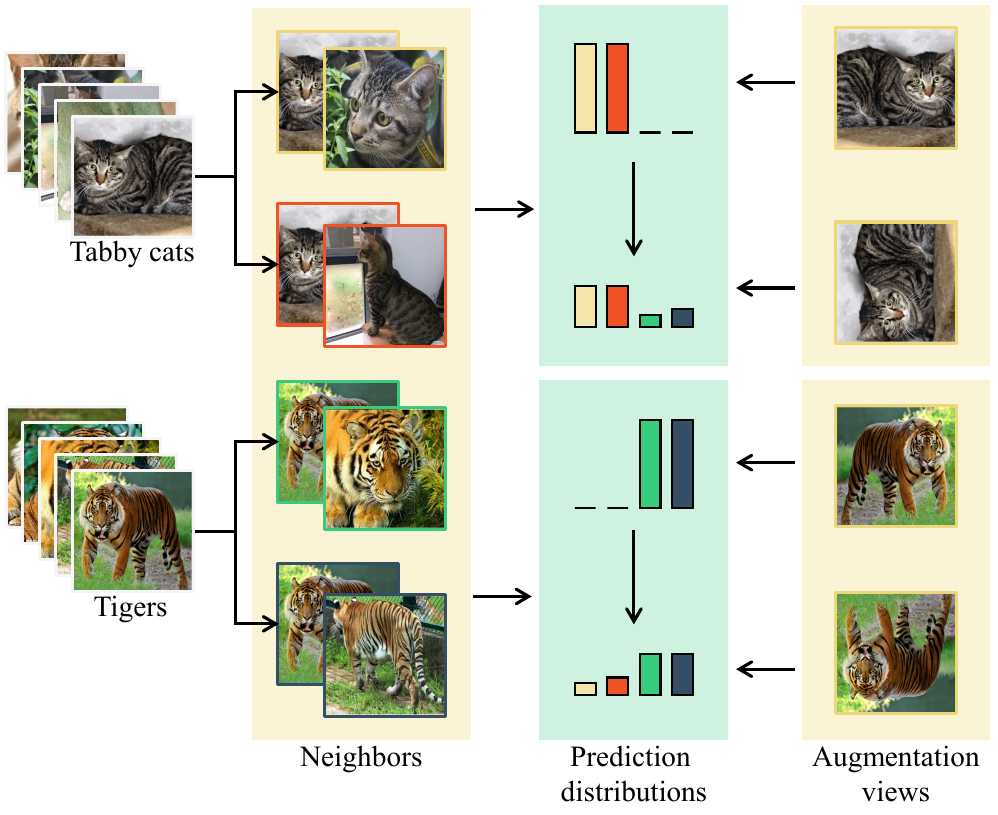}
    \caption{Each class consists of a set of local commonalities that are shared between instances within the same neighborhoods, our proposed NCRL exploits prediction distributions over these local commonalities to guide the learning of representational differences between instances of different classes. 
    }
    \label{fig:DDN_motiv}
\end{figure}

In light of this, the pioneering C-GCD method~\cite{wu2023metagcd} proposes to exploit meta-learning to learn a satisfactory initial model with less forgetting. Specifically, the proposed meta-learning strategy sets the goal of C-GCD as the meta-learning optimization objective and constructs pseudo-incremental tasks to optimize the model by mimicking the real incremental settings. However, despite achieving superior performance, it needs complicated data curation to construct the pseudo-incremental task. Importantly, it overlooks the learning of representational differences between instances of different classes which plays a key role in discovering novel classes.

Generally speaking, a good representation should possess the following two characteristics: 1) it should effectively express semantics highly relevant to its category, and 2) it should suppress semantics that are irrelevant to its category. As a result, we can obtain a discriminative feature space for the clustering/discovery of novel categories. To this end, from the perspective of methodology, 1) a line of studies~\cite{pu2023dynamic,zhao2023learning,an2023generalized} propose to incorporate clustering with contrastive learning. However, clustering algorithms are often computationally intensive or need to take the category number as a prior~\cite{an2023generalized} or other complicated and task-specific pre-processing operations~\cite{pu2023dynamic,zhao2023learning}, rendering such a solution less suitable for C-GCD. 2) Another line of studies~\cite{fini2021unified,wen2023parametric,vaze2023no} propose to exploit the self-distillation technique, wherein the sharpened prediction distribution of one augmentation view is utilized as the pseudo-label to supervise the learning of another augmentation view of the same instance. However, though clustering is not required, the prediction distributions are not meaningful enough because they are often generated by a randomly initialized classification head, which will compromise the model's performance on novel classes (see Section~\ref{dis_col}).

To address these issues, we incorporate the self-distillation technique with local commonalities and propose a novel Neighborhood Commonality-aware Representation Learning (NCRL) module. As shown in Figure~\ref{fig:DDN_motiv}, our motivation is that each class consists of a set of local similar semantics (commonalities). Meanwhile, instances within neighborhoods often share similar semantics. For instance, tabby cats exhibit analogous pointed ears and a striped pattern. These characteristics imply that we can use local commonalities derived from neighborhoods to guide the learning of representational differences between instances of different classes. Therefore, our proposed NCRL first perceives local commonalities by harnessing the average features of neighbors. Subsequently, NCoR conducts representation learning by self-distillation, where the prediction distributions are generated by exploiting the obtained local commonalities. In such a way, NCRL can generate more meaningful prediction distributions. Meanwhile, the prediction distributions, which represent the relationships between instances and semantics embodied in different classes, can help the model learn discriminative representations, thereby leading to a satisfactory clustering performance on novel classes. Furthermore, the commonality perception and representation learning are performed in a mini-batch, thus the learning process with NCRL is also efficient and not necessary to take the category number as a prior.

However, we find that only focusing on the novel class representation learning will degenerate the model's representation ability for old classes as the learning process proceeds, which in turn leads to the notorious catastrophic forgetting problem. To mitigate this issue, a natural idea is to perform knowledge distillation to maintain the learned knowledge. In general, knowledge distillation is achieved by KL divergence~\cite{zhao2022decoupled,li2023curriculum,Sun2024Logit} or MSE~\cite{Zagoruyko2017AT,liu2021exploring,guo2023class}. However, the representation-related knowledge is structured~\cite{Tian2020Contrastive}, making KL divergence or MSE have a limited effect on maintaining such knowledge. Considering the inherent advantage of contrastive learning in representing such knowledge, we further propose a Bi-level Contrastive Knowledge Distillation (BCKD) module to achieve old knowledge retention. Concretely, our proposed BCKD leverages contrastive learning to perceive both the learning and learned representational knowledge and perform knowledge distillation. By knowing what is learning and what is learned, BCKD can achieve holistic representational knowledge retention with less compromising the learned new knowledge (see Section~\ref{dis_bckd}).

Overall, taking NCRL and BCKD together, our proposed method dubbed \textbf{N}eighborhood \textbf{C}ommonality-aware \textbf{E}volution \textbf{Net}work (\textbf{NCENet}) achieves competitive performance on three C-GCD benchmark datasets. Our contributions are summarized as follows:

\begin{itemize}
    \item \textbf{A new C-GCD learning framework}. We propose a NCENet, a new C-GCD learning framework that solves the task of C-GCD from the perspective of novel class representation learning and old class representation degeneration confrontation.
    \item \textbf{Neighborhood Commonality-aware Representation Learning (NCRL) module}. NCRL incorporates local commonalities derived from neighborhoods with the self-distillation technique to guide the learning of representational differences between instances of different classes, making NCENet able to output discriminative representations for novel classes.
    \item \textbf{Bi-level Contrastive Knowledge Distillation (BCKD) module}. BCKD explores the utilization of contrastive learning in C-GCD and exploits contrastive learning to perform knowledge distillation, making NCENet could maintain the representation ability for old classes.
    \item \textbf{Competitive performance} on three C-GCD benchmark datasets.
\end{itemize}

%% file: sections/2.related_work.tex
\section{Related Work}\label{sec:related}

\subsection{Category Discovery}
\label{rel_cd}

Category Discovery (CD)~\cite{han2019learning,vaze2022generalized} aims to dynamically assign labels to unlabelled data partially based on the knowledge learned from labelled data. Contemporary research inc CD can be roughly divided into two groups, Novel Category Discovery (NCD)~\cite{han2021autonovel,jia2021joint,zhong2021neighborhood} and Generalized Category Discovery (GCD)~\cite{chiaroni2023parametric,rastegar2023learn,pu2023dynamic}. NCD operates under the premise that the label space of the unlabeled data is entirely separate from that of the labeled data. In contrast, GCD generalizes the NCD by considering a scenario where the unlabeled data encompasses known and previously unseen classes. Despite the differences between the two tasks, one of the key challenges is representation learning. In light of this, supervised contrastive learning~\cite{khosla2020supervised} and unsupervised contrastive learning~\cite{vaze2022generalized} serve as a baseline solution. To further enhance representation learning, recent studies can be roughly grouped into neighborhood-based, clustering-based, and self-distillation methods.
Considering that the number of negative samples in unsupervised contrastive learning is dominant, it will undermine the performance of representation learning. The neighborhood-based methods~\cite{zhang2023promptcal,zhong2021neighborhood,Choi_2024_CVPR} introduce neighbors to mitigate this issue. For example, NCL~\cite{zhong2021neighborhood} utilizes instances within the neighborhood of the anchor sample as positive samples and mines hard negative samples from a memory buffer, while CMS~\cite{Choi_2024_CVPR} leverages mean-shifted embeddings derived from neighborhoods as contrastive samples. Unlike the neighborhood-based methods, the clustering-based methods~\cite{an2023generalized,zhao2023learning,pu2023dynamic} argue that unsupervised contrastive learning can not underline relationships between instances of the same classes. To address this issue, the clustering-based methods leverage various clustering algorithms, such as GMM~\cite{zhao2023learning} or Infomap~\cite{pu2023dynamic}, and prototypical contrastive learning to learn representation. In contrast to the clustering-based methods, the self-distillation-based methods~\cite{fini2021unified,wen2023parametric,vaze2023no,wang2024sptnet} perform representation learning by minimizing the prediction distributions of two augmentation view of the same instance, where a random initialized classification head is used to generate prediction distributions.

In contrast to these offline methods, our proposed method focuses on solving both the representation learning of sequential unlabelled data and the catastrophic forgetting problem that occurs in the continuous learning process. More concretely, our proposed NCRL is most similar to self-distillation-based methods, but our NCRL exploits commonalities derived from neighborhoods to output more meaningful prediction distributions.

\subsection{Incremental Category Discovery}

Incremental Category Discovery (ICD) aims to continuously discover novel classes from unlabelled data while maintaining the ability for old classes. Recent studies~\cite{roy2022class,joseph2022novel,zhang2022grow,zhao2023incremental,wu2023metagcd} mainly engage in four ICD tasks, class-incremental Novel Class Discovery (class-iNCD)~\cite{roy2022class,joseph2022novel}, Continuous Category Discovery (CCD)~\cite{zhang2022grow}, Incremental Generalized Category Discovery (IGCD) and Continuous Generalized Category Discovery (C-GCD). In these tasks, class-iNCD and CCD mainly focus on the incremental learning of NCD, while IGCD and C-GCD focus on the incremental learning of GCD. Further, class-iNCD only sets one incremental stage while CCD sets multiple incremental stages. Meanwhile, except for C-GCD, the other tasks utilize the same data to train and evaluate the model. To address the task of class-iNCD, FRoST~\cite{roy2022class} retrains the prototypes of labelled data and replays them in incremental sessions followed by a feature-level distillation loss to prevent the forgetting problem. ADM~\cite{chen2024adaptive} sets a base branch to maintain the previously learned knowledge and a novel branch to discover novel classes. At the end of each learning session, ADM merges the two branches with an adaptive module to prevent the growth of the model's parameters. To solve the task of CCD, GM~\cite{zhang2022grow} presents a learning framework consisting of a growing phase and a merging phase. In the growing phase, GM first detects novel samples and then sets an additional dynamic branch to perform the NCD task with the detected novel samples and previously learned static branch. In the merging phase, GM first learns class-level discriminative features and then merges the two branches in an EMA manner. For the task of IGCD, Zhao \etal ~\cite{zhao2023incremental} provide a baseline by adapting the SimGCD to this task. For the challenging task of C-GCD, MetaGCD~\cite{wu2023metagcd} introduces a meta-learning framework that solves C-GCD from the perspective of model initialization.

In this paper, we focus on solving the challenging task of C-GCD. Unlike MetaGCD, we solve C-GCD from the perspective of representation learning. More concretely, our proposed method leverages local commonality derived from neighborhood to learn representations for novel classes and contrastive learning to mitigate the representation degeneration of old classes.

\subsection{Knowledge distillation}

Knowledge distillation~\cite{hinton2015distilling} aims to transfer ``Dark Knowledge'' from a larger model (teacher) to a smaller model (student). The existing KD methods can be roughly divided into two groups, logits distillation and feature distillation.

In logits distillation, TAKD~\cite{mirzadeh2020improved} introduces several teacher assistants with a gradual reduction of model size to achieve progressive knowledge transfer. DGKD~\cite{son2021densely} improves TAKD by gathering logits of previous teacher assistants. DIST~\cite{huang2022knowledge} proposes to use the Pearson correlation coefficient~\cite{pearson1896vii} derived from logits to match the inter- and intra-correlations between teacher and student. GLD~\cite{kim2021distilling} proposes to add an additional local logits distillation branch to further transfer spatial knowledge.
DKD~\cite{zhao2022decoupled} splits logits into the target and non-target parts and performs knowledge distillation in a decoupled manner. LSDK~\cite{Sun2024Logit} proposes a logits standardization method to help the student model capture key information of the teacher model. CTKD~\cite{li2023curriculum} sets the knowledge distillation temperature to be trainable and proposes a learning curriculum to control the difficulty of learning tasks.

In feature distillation, a line of works engage in the designation of various feature-oriented distillation knowledge, such as intermediate features~\cite{adriana2015fitnets,heo2019comprehensive}, cross-layer fusion features~\cite{chen2021distilling}, relationships~\cite{liu2019knowledge,tung2019similarity,peng2019correlation} between instances, attention maps~\cite{Zagoruyko2017AT,liu2021exploring,guo2023class}. In the above methods, KL divergence and MSE are usually used to perform knowledge distillation. In contrast to these methods, CRD~\cite{Tian2020Contrastive} argues that representational knowledge is structured and proposes to leverage contrastive learning to achieve representational knowledge transfer, where a memory buffer is set to store negative samples. 

Inspired by CRD~\cite{Tian2020Contrastive}, our proposed method leverages the contrastive learning technique to conquer the representation degeneration issue. But unlike CRD, our proposed BCKD performs KD in a bi-level contrastive manner to achieve comprehensive knowledge retention.

\subsection{Representation Learning with self-distillation}
In addition to the methods introduced in Section~\ref{rel_cd}, a line of works in Semi-Supervised Learning~\cite{assran2021semi,berthelot2019mixmatch} and Self-Supervised Learning~\cite{grill2020bootstrap,caron2021emerging,fang2021seed,assran2022masked} also utilize self-distillation for representation learning. When it comes to generating prediction distributions, these methods can be categorized into two types: instance-based~\cite{fang2021seed,assran2021semi} and prototype-based~\cite{caron2020unsupervised,caron2021emerging,assran2022masked}. Instance-based methods either use labeled support instances~\cite{assran2021semi} from sampled classes or random instances~\cite{fang2021seed} to produce predictions. In prototype-based methods, the prototypes are typically set to be trainable. Unlike these methods, our proposed NCRL uses local commonalities derived from instances within different neighborhoods to generate prediction distributions.

\begin{figure*}
    \centering
    \includegraphics[width=1.99\columnwidth]{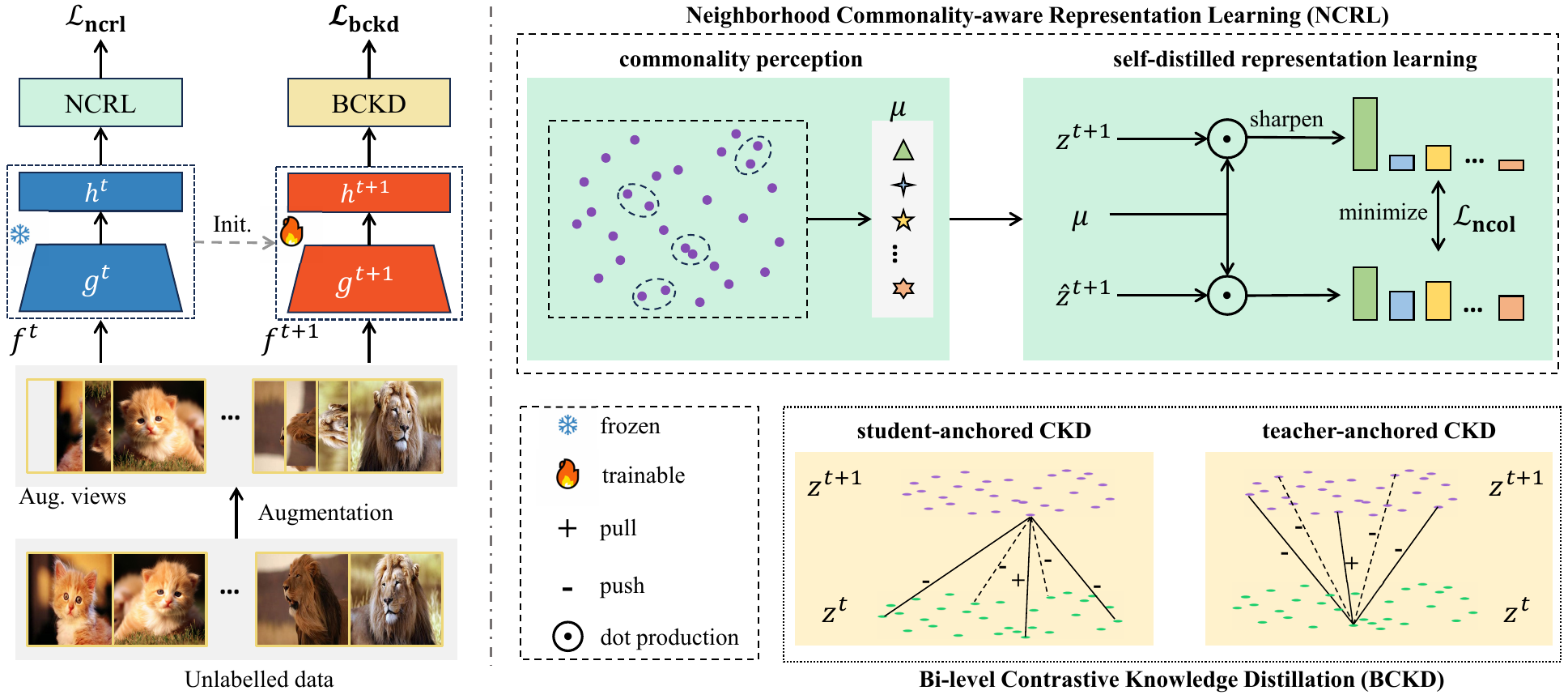}
    \caption{Pipeline of our proposed incremental learning framework. Our proposed method leverages the Neighborhood Commonality-aware Representation Learning (NCRL) module to learn representations for novel classes and the Bi-level Contrastive Knowledge Distillation (BCKD) module to maintain the representation ability for old classes. In NCRL, local commonalities $\mu$ derived from neighborhoods are used to generate prediction distributions, and a self-distillation technique is used to learn representations. In BCKD, student-anchored contrastive knowledge distillation and teacher-anchored contrastive knowledge distillation are performed to achieve holistic representational knowledge retention.}
    \label{fig:DDN_method}
\end{figure*}

%% file: sections/3.problem_formulation.tex
\section{Preliminaries}\label{sec:problem}

\noindent\textbf{Task Definition.} 
In C-GCD, a base session and several incremental sessions come in sequence. The base session provides sufficient labelled data, whereas the incremental sessions only provide unlabelled data. The goal of C-GCD is to continually discover novel classes without forgetting old classes. Formally, let $\mathcal{D}^0$ denotes the base session and $\mathcal{D}^t(t\textgreater 0)$ denotes the incremental session. The label spaces of different sessions satisfy $\mathcal{Y}^{t-1}\subset \mathcal{Y}^t$, which means that data of incremental session $t$ comes from seen and unseen categories. The training data of $\mathcal{D}^0$ and $\mathcal{D}^t(t>0)$ satisfy  $\mathcal{D}^0_\text{train}{\cap}\mathcal{D}^t_\text{train}=\emptyset$. In incremental learning session $t$, only $\mathcal{D}^t_\text{train}$ is available. When finishing the training, the model is evaluated with test data accumulated until session $t$, \ie, the test set $\mathcal{D}^t_\text{test}$ of incremental session $t$ is constituted by $\{\mathcal{D}^0_\text{test},...,\mathcal{D}^t_\text{test}\}$.

\noindent\textbf{Architecture.} 
Following~\cite{wu2023metagcd}, our model architecture $f={g}\circ{h}$ consists of an encoder $g$ and a projection head $h$. In the training stage, our goal is to optimize parameters of $f$ using provided training data. In the inference stage, we use $g$ to encode corresponding test data and clustering accuracy on encoded features to evaluate the model's performance.

\noindent\textbf{Learning Startup.} In the base session, we follow the common practice~\cite{vaze2022generalized,wen2023parametric,pu2023dynamic} to combine supervised and unsupervised contrastive learning to train the model. Formally, let $z_i$ and $\hat{z}_i$ denote projected features obtained by passing two augmentation views of the same instance into $f$. The supervised contrastive loss $\mathcal{L}_\text{sup}$ is calculated by: 
\begin{equation}
    \mathcal{L}_\text{sup}=\frac{1}{|B^l|}\sum_{i}\frac{1}{|\mathcal{N}(i)|}\sum_{q\in{\mathcal{N}(i)}}-\text{log}\frac{{\exp(z_i \cdot z_q/{\tau}_{{r}})}}{\sum_{j\neq{i}}{\exp(z_i \cdot z_j/{\tau}_{{r}})}},
\end{equation}
where $|B^l|$ denotes the number of labeled data in a mini-batch, $\mathcal{N}(i)$ denotes indices of other instances with the same label as instance $i$ and ${\tau}_{{r}}$ is the scaling factor.
The unsupervised contrastive loss $\mathcal{L}_\text{unsup}$ is defined as:
\begin{equation}
\label{eq:unsup}
    \mathcal{L}_\text{unsup}=\frac{1}{|B|}\sum-\text{log}\frac{{\exp(z_i \cdot \hat{z}_i/{\tau}_{{r}})}}{\sum_{j\neq{i}}{\exp(z_i \cdot z_j/{\tau}_{{r}})}},
\end{equation}
where $|B|$ denotes the batch size.
After obtaining $\mathcal{L}_\text{sup}$ and $\mathcal{L}_\text{unsup}$, the overall objective in the base session is represented as:
\begin{equation}
    \mathcal{L} = {\beta} \mathcal{L}_\text{sup} + (1-{\beta}) \mathcal{L}_\text{unsup}
\end{equation}
where ${\beta}$ is a hyperparameter used to control the contribution of $\mathcal{L}_\text{sup}$ and $\mathcal{L}_\text{unsup}$.

%% file: sections/4.method.tex
\section{Methodology}\label{sec:method}

\subsection{Overview}

As depicted in Figure~\ref{fig:DDN_method}, our proposed NCENet comprises two key components: the Neighborhood Commonality-aware Representation Learning (NCRL) module (Section~\ref{sec:csdl}) and the Bi-level Contrastive Knowledge Distillation (BCKD) module (Section~\ref{sec:bckd}). The NCRL module is primarily responsible for discriminative representation learning of novel classes, while BCKD is mainly designed to preserve the old representational knowledge. Concretely, for the incremental session $t+1$, NCENet commences the incremental learning process by generating two augmentation views for each unlabelled instance in a mini-batch. Then, NCENet feeds different augmentation views into the current learning model $f^{t+1}$ and a frozen historical model $f^{t}$ obtained from the last session. Here, we denote corresponding outputs as $z^{t+1}$ and $z^{t}$. Next, NCENet inputs $z^{t+1}$ into NCRL to perform representation learning. Simultaneously, NCENet feeds $z^{t+1}$ and $z^{t}$ into BCKD to conduct knowledge distillation. Let $\mathcal{L}_\text{ncrl}$ and $\mathcal{L}_\text{bckd}$ denote the learning objectives of NCRL and BCKD, respectively, the overall learning objective of NCENet is defined as:
\begin{equation}
    \mathcal{L} = {\lambda}_b \mathcal{L}_\text{ncrl} + (1-{\lambda}_b) \mathcal{L}_\text{bckd},
\end{equation}
where ${\lambda}_b$ refers to a hyperparameter used to balance the contributions of NCRL and BCKD.

\subsection{Neighborhood Commonality-aware Representation Learning }
\label{sec:csdl}

The core idea of the designation of NCRL is to exploit local commonalities derived from instances within different neighborhoods to guide the learning of representational differences between instances of different classes. To this end, NCRL mainly involves two steps: 1) commonality perception and 2) self-distilled representation learning.

{\bf Step1: commonality perception} used to obtain local commonalities to prepare for future commonality learning. Concretely, given features $z^{t+1}\in\mathbb{R}^{|B|\times{d}}$ encoded by current learning model $f^{t+1}$, where $d$ refers to the feature dimension. NCRL first calculates cosine similarities ${\omega}\in\mathbb{R}^{|B|\times{|B|}}$ between different features. Then, NCRL selects $k$ nearest neighbors $NN(z^{t+1}_i)\in\mathbb{R}^{k\times{d}}$ for each feature based on obtained $\omega$. In the end, NCRL computes the local commonalities $\mu\in{\in\mathbb{R}^{|B|\times{d}}}$ by:
\begin{equation}
    \mu_i = \frac{1}{k}\sum_{q\in NN(z^{t+1}_i)}z^{t+1}_q,
\end{equation}
where $\mu_i$ denotes the local commonality derived from neighbors of $z^{t+1}_i$.

{\bf Step2: self-distilled representation learning} leverages obtained local commonalities to learn discriminative representations for novel classes. Concretely, given features $z^{t+1}$ and $\hat{z}^{t+1}$ of two augmentation views of the same instance. NCRL first computes prediction distribution $p$ of $z^{t+1}$ over $\mu$ by:
\begin{equation}
\label{eq:prob}
    p_i^j = \frac{\exp(z^{t+1}_i\cdot{\mu}_j/{\tau})}{\sum_m\exp(z^{t+1}_i\cdot{\mu}_m/{\tau})},
\end{equation}
where $p_i^j$ refers to the probability of $z^{t+1}_i$ belonging to local commonality ${\mu}_j$ and $\tau$ refers to temperature used to sharpen the prediction distribution. Meanwhile, using Eq.~\ref{eq:prob} and setting $\tau$ to 1, NCRL computes prediction distribution $\hat{p}$ of $\hat{z}^{t+1}$ over $\mu$. After obtaining $p$ and $\hat{p}$, the learning objective of NCRL is defined as:
\begin{equation}
    \mathcal{L}_{\text{ncrl}}=-\frac{1}{|B|}\sum_{i=1}^{|B|}\sum_j{p_i^j\log{\hat{p}_i^j}}.
\end{equation}

\noindent{\textbf{Remark:}} Though the local commonalities obtained from a mini-batch are not comprehensive, a wider and more diverse set of instances will compensate for this shortcoming as the learning process proceeds.

\subsection{Bi-level Contrastive Knowledge Distillation}
\label{sec:bckd}
With NCRL, we can improve the model's representation ability for novel classes. However, the absence of old training data will degenerate the model's representation ability for old classes as the learning process proceeds, this phenomenon is also dubbed the dilemma of plasticity and stability~\cite{LIU2024111612,lwf}. To achieve old representational knowledge retention, an effective solution is to apply the contrastive learning-based knowledge distillation method~\cite{Tian2020Contrastive} used for the model compression task to C-GCD. However, the differences between C-GCD and model compression tasks will make such a method suffer from the over-constraint issue. Specifically, in C-GCD, we expect the student model to inherit knowledge from the teacher model without hindering the learning of new knowledge. In model compression tasks, more emphasis is placed on the student model's ability to fully inherit all knowledge from the teacher model. Consequently, directly using existing contrastive knowledge distillation compression methods in C-GCD may result in the student model being overly reliant on the teacher's knowledge, which may undermine the learning of new knowledge to some extent (Section\ref{dis_bckd}).

In light of this, BCKD leverages student-anchored contrastive knowledge distillation and teacher-anchored contrastive knowledge to achieve representational knowledge transportation from teacher to student. Formally, given features $z^{t+1}$ encoded by current learning model $f^{t+1}$ and $z^{t}$ encoded by historical model $f^{t}$. The student-anchored contrastive knowledge distillation learning objective is defined as:
\begin{equation}
    \mathcal{L}_{\text{sa}}= -\frac{1}{|B|}\sum_{j}\log\frac{\exp(z^{t+1}_j \cdot z^{t}_j/{\tau}_\text{k})}{\sum_{i}\exp(z^{t+1}_j \cdot z^{t}_i/{\tau}_\text{k})},
\end{equation}
where ${\tau}_\text{k}$ refers to the temperature. The teacher-anchored contrastive knowledge distillation learning objective is defined as
\begin{equation}
    \mathcal{L}_{\text{ta}}= -\frac{1}{|B|}\sum_{j}\log\frac{\exp(z^{t}_j \cdot z^{t+1}_j/{\tau}_\text{k})}{\sum_{i}\exp(z^{t}_j \cdot z^{t+1}_i/{\tau}_\text{k})}.
\end{equation}
Overall, the learning objective of BCKD is presented as:
\begin{equation}
    \mathcal{L}_\text{bckd} = \frac{\mathcal{L}_{\text{sa}}+\mathcal{L}_{\text{ta}}}{2}.
\end{equation}

By incorporating student-anchored contrastive knowledge distillation with teacher-anchored contrastive knowledge, BCKD can perceive the learning and learned representational knowledge, thus achieving effective incremental-oriented representational knowledge retention. Additionally, since knowledge distillation is conducted within a mini-batch, BCKD obviates the need for a memory buffer to store negative samples.

%% file: sections/5.experiment.tex
\section{Experiments}\label{sec:exp}

\subsection{Datasets}

\begin{table}[ht]
\centering
\caption{ Statistics of each dataset used in our experiments.}
\label{tab:dataset}
\setlength{\tabcolsep}{.9mm}{
\begin{tabular}{l|cc|cc|c}
\toprule[1pt]
\multirow{2}{*}{Dataset} &\multicolumn{2}{c|}{Labelled Set} &\multicolumn{2}{c|}{Unlabelled Set} & \multirow{2}{*}{\#Session}\\ 
\cline{2-5}
                & \#class   &\#image    &\#class    &\#image & \\\hline
CIFAR10         & 7         & 28000     & 10        & 22000  & 4 \\
CIFAR100        & 80        & 32000     & 100       & 18000  & 5 \\
Tiny-ImageNet   & 150       & 60000     & 200       & 40000  & 6 \\
\bottomrule[1pt]
\end{tabular}
}
\end{table}

We conduct corresponding experiments on three benchmark datasets, including CIFAR10~\cite{krizhevsky2009learning}, CIFAR100~\cite{krizhevsky2009learning} and Tiny-ImageNet~\cite{Le2015TinyIV}. Following~\cite{wu2023metagcd}, we split CIFAR10 dataset into 1 base session and 3 incremental sessions. For the CIFAR100, a division is made into 1 base session and 4 incremental sessions. In the case of Tiny-ImageNet, it is structured into 1 base session and 5 incremental sessions. For each dataset, we sample 80\% training images from each labelled class for base learning, the remaining data are used for incremental learning. We summarize dataset splits in Table~\ref{tab:dataset}.

\noindent \textbf{Incremental session.} For CIFAR10, the training data of each incremental session incorporates 3,000 training images from 1 novel class and 2,000 training images from $7+(t-1) \times 1$ seen classes, where $t$ refers to the incremental session id. For CIFAR100, 1,500 training images from 5 novel classes and 2,000 training images from $80+(t-1) \times 5$ seen classes are used for incremental learning. For Tiny-ImageNet, we sample 3,000 training images from 10 novel classes and 3,000 training images from $150+(t-1) \times 10$ known classes to construct the training data.

\begin{table*}[ht]\small
    \centering
    \caption{Performance (in \%) comparisons with other methods on CIFAR10, CIFAR100, and Tiny-ImageNet. The performance of other methods are provided by~\cite{wu2023metagcd}. Our proposed method shows consistent superiority over other methods on \texttt{New} classes.}
    \label{tab:NCENet_cmp_sota}
    \setlength{\tabcolsep}{2.2mm}{
    \begin{tabular}{lccc|ccc|ccc|ccc}
    \toprule[1pt]
    \multirow{3}{*}{Methods} & \multicolumn{9}{c}{CIFAR10 (Session Number)} & \multicolumn{3}{c}{Final} \\ 
    \cline{2-10}                       
    & \multicolumn{3}{c}{1} & \multicolumn{3}{c}{2} & \multicolumn{3}{c}{3} & \multicolumn{3}{c}{Impro.}  \\ 
    \cline{2-13} 
    & All & Old & New & All & Old & New & All & Old & New & All & Old & New\\
    \hline
    RankStats\cite{han2021autonovel}    & 69.31 & 70.20 & 58.63 & 65.23 & 67.86 & 51.20 & 38.16 & 50.01 & 35.94 & +55.67 & +44.62 & +56.03
 \\
    FRoST\cite{roy2022class}            & 73.92 & 81.17 & 66.45 & 69.56 & 79.73 & 58.04 & 67.73 & 70.84 & 51.13 & +26.10 & +23.79 & +40.84
 \\
    VanillaGCD\cite{vaze2022generalized}& 89.24 & 97.97 & 81.80 & 85.13 & 96.67 & 74.60 & 86.41 & 95.03 & 76.75 & +7.42 & -0.40 & +15.22
 \\
    GM\cite{zhang2022grow}              & 90.00 & 98.41 & 77.40 & 87.39 & 99.01 & 73.46 & 87.86 & 97.15 & 78.93 & +5.97 & -2.52 & +13.04
 \\
    MetaGCD\cite{wu2023metagcd}         & 95.38 & 99.07 & 89.15 & 93.34 & 98.81 & 85.39 & 92.66 & 97.23 & 84.71 & +1.17 & -2.60 & +7.26
 \\
    \hline
    NCENet(Ours)                           & 96.13 & 96.96 & 90.30 & 93.76 & 96.03 & 85.80 & 93.83 & 94.63 & 91.97 & {} & {} & {} \\ 
    \bottomrule[1pt]
    \\
    \end{tabular}}
   
    \setlength{\tabcolsep}{1.11mm}{
    \begin{tabular}{lccc|ccc|ccc|ccc|ccc}
    \toprule[1pt]
    \multirow{3}{*}{Methods} &\multicolumn{12}{c}{CIFAR100 (Session Number)} &\multicolumn{3}{c}{Final} \\ 
    \cline{2-13}                         
    &\multicolumn{3}{c}{1} &\multicolumn{3}{c}{2} &\multicolumn{3}{c}{3} &\multicolumn{3}{c}{4} &\multicolumn{3}{c}{Impro.}  \\ 
    \cline{2-16} 
    & All & Old & New & All & Old & New & All & Old & New & All & Old & New & All & Old & New\\
    \hline
    RankStats\cite{han2021autonovel}    & 62.33 & 64.22 & 31.60 & 55.01 & 58.55 & 26.85 & 51.77 & 56.70 & 25.47 & 47.51 & 54.59 & 17.20 & +28.37 & +26.10 & +50.25 \\
    FRoST\cite{roy2022class}            & 67.14 & 68.57 & 50.73 & 67.01 & 68.82 & 52.60 & 62.35 & 65.48 & 45.67 & 55.84 & 59.06 & 42.95 & +20.04 & +21.63 & +24.50 \\
    VanillaGCD\cite{vaze2022generalized}& 76.78 & 77.91 & 58.60 & 73.67 & 75.29 & 60.70 & 72.77 & 74.72 & 62.33 & 71.44 & 74.75 & 58.20 & +4.45 & +5.94 & +6.95 \\
    GM\cite{zhang2022grow}              & 78.29 & 79.91 & 66.00 & 77.58 & 79.64 & 61.13 & 74.56 & 77.60 & 58.14 & 72.02 & 75.98 & 56.32 & +3.86 & +4.71 & +11.13 \\ 
    MetaGCD\cite{wu2023metagcd}         & 78.96 & 79.36 & 72.60 & 78.67 & 79.41 & 66.81 & 76.06 & 78.20 & 64.87 & 74.56 & 77.60 & 61.13 & +1.32 & +3.09 & +6.32 \\ 
    \hline
    NCENet(Ours)                           & 80.85 & 82.61 & 70.40 & 78.97 & 81.68 & 69.90 & 77.41 & 81.48 & 72.27 & 75.88 & 80.69 & 67.45
 & {} & {} & {} \\ 
    \bottomrule[1pt]
    \\
    \end{tabular}}

    \setlength{\tabcolsep}{.36mm}{
    \begin{tabular}{lccc|ccc|ccc|ccc|ccc|ccc}
    \toprule[1pt]
    \multirow{3}{*}{Methods} & \multicolumn{15}{c}{Tiny-ImageNet (Session Number)} &\multicolumn{3}{c}{Final} \\ 
    \cline{2-16}                     
    & \multicolumn{3}{c}{1} & \multicolumn{3}{c}{2} &\multicolumn{3}{c}{3} &\multicolumn{3}{c}{4} &\multicolumn{3}{c}{5} &\multicolumn{3}{c}{Impro.}  \\ 
    \cline{2-19} 
    & All & Old & New & All & Old & New & All & Old & New & All & Old & New & All & Old & New & All & Old & New\\
    \hline
    RankStats\cite{han2021autonovel}    & 62.39 & 64.54 & 35.01 & 55.89 & 52.23 & 34.20 & 49.88 & 46.17 & 28.33 & 44.20 & 42.87 & 24.50 & 36.09 & 35.20 & 15.76 & +36.73 & +41.25 & +46.48  \\ 
    FRoST\cite{roy2022class}            & 64.92 & 67.84 & 46.28 & 59.50 & 61.86 & 40.60 & 57.86 & 60.63 & 39.14 & 55.68 & 59.71 & 36.55 & 50.49 & 53.76 & 33.37 & +22.33 & +22.69 & +28.87  \\ 
    VanillaGCD\cite{vaze2022generalized}& 75.92 & 78.17 & 62.15 & 74.53 & 77.73 & 56.12 & 73.64 & 74.85 & 57.31 & 70.69 & 71.13 & 54.35 & 66.15 & 67.17 & 54.43 & +6.67 & +9.28 & +7.81  \\ 
    GM\cite{zhang2022grow}              & 76.32 & 79.55 & 63.60 & 75.43 & 78.10 & 57.40 & 72.63 & 76.29 & 54.80 & 70.54 & 76.80 & 51.50 & 67.31 &72.08 & 50.90 & +5.51 & +4.37 & +11.34  \\  
    MetaGCD\cite{wu2023metagcd}         & 78.67 & 79.41 & 66.80 & 77.89 & 79.95 & 61.40 & 75.23 & 77.86 & 61.20 & 72.00 & 75.61 & 57.55 & 70.24 & 71.53 & 58.46  & +2.58 & +4.92 & +3.78  \\  
    \hline
    NCENet(Ours)                           & 77.14 & 78.13 & 67.20 & 76.58 & 78.51 & 65.20 & 74.79 & 78.53 & 64.00 & 72.94 & 77.01 & 61.20	 & 72.82 & 76.45 & 62.24 & {} & {} & {}  \\ 
    \bottomrule[1pt]
    \end{tabular}}
    
\end{table*}

\subsection{Evaluation Protocol}

After finishing the training in each incremental session, we follow~\cite{wu2023metagcd} to measure the clustering accuracy ($ACC$) by
\begin{equation}
\label{eq:eval}
    ACC=\frac{1}{M}\sum_{i=1}^{M}\mathbb{I}\{y^*_i=m\hat{y}_i\},
\end{equation}
where $M$ refers to the total number of test images used in the current session, $y^*$ indicates the ground truth, $\hat{y}$ represents the cluster label given by our model and  $m$ refers to the optimal permutation for matching predicted cluster assignment to the ground truth, and $\mathbb{I}$ denotes the indicator function. In this paper, we use clustering accuracy on \textit{All} classes to evaluate the model's entire performance. To decouple the evaluation on $forgetting$ and $discovery$, we follow~\cite{wu2023metagcd} to further report clustering accuracy on \textit{Old} classes and \textit{New} classes. Concretely, when computing the clustering accuracy on \textit{Old}/\textit{New} classes, we only use samples in the test set belonging to \textit{Old}/\textit{New} classes.

\subsection{Implementation Details}

We use PyTorch~\cite{paszke2019pytorch} to implement our proposed method and conduct all experiments using one NVIDIA GeForce RTX 2080 Ti.

\noindent \textbf{Model Architecture.} Follow~\cite{wu2023metagcd}, we adopt ViT-B/16 pretrained by DINO~\cite{caron2021emerging} as the encoder and take the encoder's output \texttt{[CLS]} token with a dimension of 768 as the feature representation. We build the projection head using three linear layers, where we set the hidden dimension to 2048 and the output dimension to 65536 as~\cite{wu2023metagcd}. In following training processes, we only finetune the last block of the encoder and the projection head.

\noindent \textbf{Base Training.} We split the provided labelled set into a training set and a validation set used to select the best model. In particular, the training set takes 75\% samples, and the validation set takes the remaining  25\% samples. We train the model with a batch size of 128 for 50 epochs. We adopt SGD as the optimizer, where the initial learning rate is set to 0.01. We decay the learning rate with the cosine schedule~\cite{loshchilov2017sgdr}. We set ${\tau}_r$ to 0.1 and ${\beta}$ to 0.35 as~\cite{wen2023parametric,vaze2022generalized}.

\noindent \textbf{Incremental Training.} We train the model with a batch size of 128 for 20 epochs. We adopt SGD as the optimizer, where the initial learning rate is set to 0.0001 and decayed using the cosine schedule~\cite{loshchilov2017sgdr}. We set the temperature ${\tau}$ in NCRL, temperature ${\tau}_k$ in BCKD, and hyperparameter ${\lambda}_b$ used to control the contributions of the two modules to 0.07, 0.04, and 0.1, respectively.

\begin{table*}[ht]
\begin{center}
\caption{{\bf Ablation study of various components of our NCENet on the CIFAR100 dataset}. We report \textit{All}/\textit{Old} /\textit{New} class accuracy for each incremental session, and the average of all sessions such as mean \textit{All} ($mA$), mean \textit{Old} ($mO$) and mean \textit{New} accuracy ($mN$).}
\label{tab:ablat_NCENet}
\setlength{\tabcolsep}{2.2mm}{
\begin{tabular}{l|ccc|ccc|ccc|ccc}
\toprule[1pt]
\multirow{3}{*}{Methods} &\multicolumn{9}{c|}{Session Number}  &\multicolumn{3}{c}{Average} \\ \cline{2-10}
                         &\multicolumn{3}{c|}{1} &\multicolumn{3}{c|}{2} &\multicolumn{3}{c|}{3}  &\multicolumn{3}{c}{Acc}\\ 
                         & All & Old & New & All & Old & New & All & Old & New  & $m$A & $m$O & $m$N \\ \hline
w/o $\mathcal{L}_\text{ncrl}$     & 95.58 & 97.10 & 84.90 & 91.76 & 96.61 & 74.75 & 90.34 & 96.20 & 76.67 & 92.56 & 96.64 & 78.77  \\
w/o $\mathcal{L}_\text{bckd}$    & 95.34 & 94.89 & 98.50 & 91.00 & 89.21 & 97.25 & 89.46 & 86.61 & 96.10 & 91.93 & 90.24 & 97.28 \\
\midrule
w all                            & 96.13 & 96.96 & 90.30 & 93.76 & 96.03 & 85.80 & 93.83 & 94.63 & 91.97 & 94.57 & 95.87 & 89.36 \\
\bottomrule[1pt]
\end{tabular}}
\end{center}
\end{table*}

\begin{figure*}
    \centering
    \includegraphics[width=1.9\columnwidth]{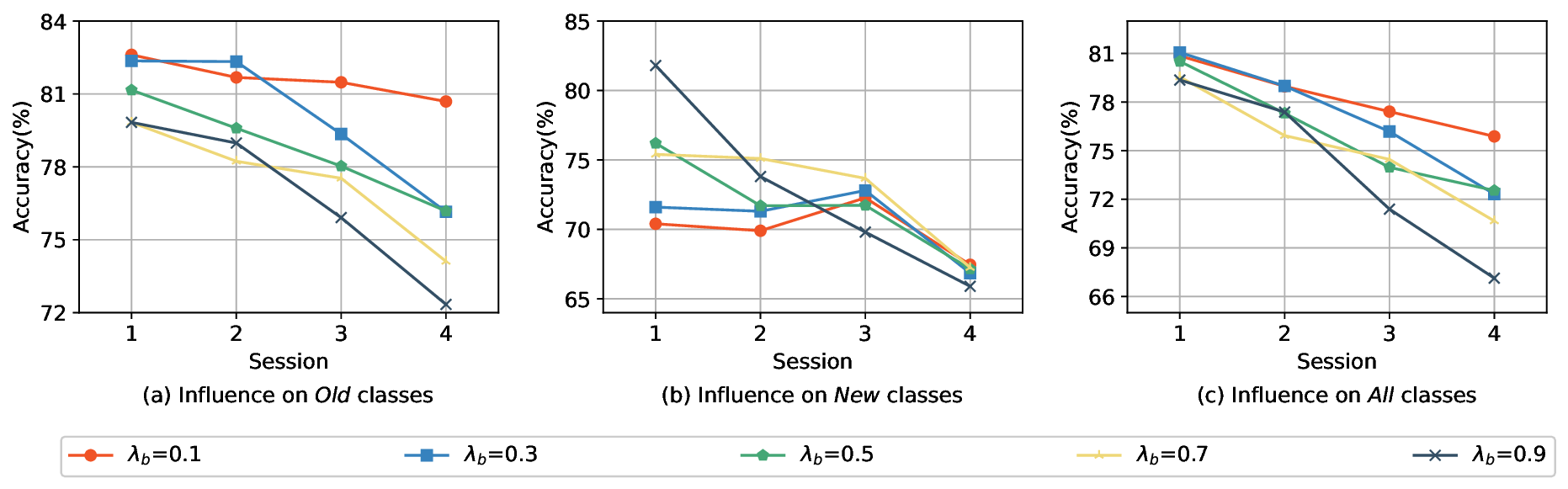}
    \caption{Clustering accuracy in each incremental learning session under different balance factor ${\lambda}_b$. Our proposed method prefers a small balance factor. }
    \label{fig:NCENet_plast}
\end{figure*}

\subsection{Comparison with State-of-the-Art}

To validate the effectiveness of our proposed NCENet, we compare NCENet with the novel category discovery method (FRoST\cite{roy2022class}), generalized category discovery method (VanillaGCD\cite{vaze2022generalized}), incremental category discovery methods (FRoST\cite{roy2022class} and GM\cite{zhang2022grow}), and a strong C-GCD baseline (MetaGCD~\cite{wu2023metagcd}). 

Table~\ref{tab:NCENet_cmp_sota} shows the clustering accuracy on \textit{Old}/\textit{New}/\textit{All} of each method in each session. On CIFAR10, most methods achieve superior performance. Especially, the previous state-of-the-art method MetaGCD establishes a strong baseline. Compared to MetaGCD, though our proposed NCENet shows no advantage on \textit{Old} class, NCENet achieves better clustering performance on \textit{New} and \textit{All} classes in each incremental session. Particularly, the clustering accuracy on \textit{New} classes of NCENet surpasses that of MetaGCD by a large margin of 7.26\%.

On CIAFR100, the clustering accuracy on \textit{Old} and \textit{All} classes of our proposed NCENet shows consistent superiority over other methods. Further, though the clustering accuracy on \textit{New} classes of NCENet is weaker than the second-best method MetaGCD in the first incremental session, NCENet outperforms MetaGCD in the last three incremental sessions. Particularly, in the last incremental session, NCENet outperforms MetaGCD by a margin of 3.09\%, 6.32\%, and 1.32\% on \textit{Old}, \textit{New} and \textit{All} classes, respectively.

On Tiny-ImageNet, our proposed NCENet outperforms other methods on \textit{New} classes in each incremental session. As for the performance on \textit{Old} classes, compared to MetaGCD, though NCENet shows less competitive performance in the first two sessions, NCENet achieves better performance in the last three incremental sessions.  Particularly, in the last session, compared to MetaGCD, NCENet achieves an improvement of 2.58\%, 4.92\%, and 3.78\% on \textit{Old} classes, \textit{New} classes, and \textit{All} classes, respectively.

\begin{figure*}
    \centering
    \includegraphics[width=1.9\columnwidth]{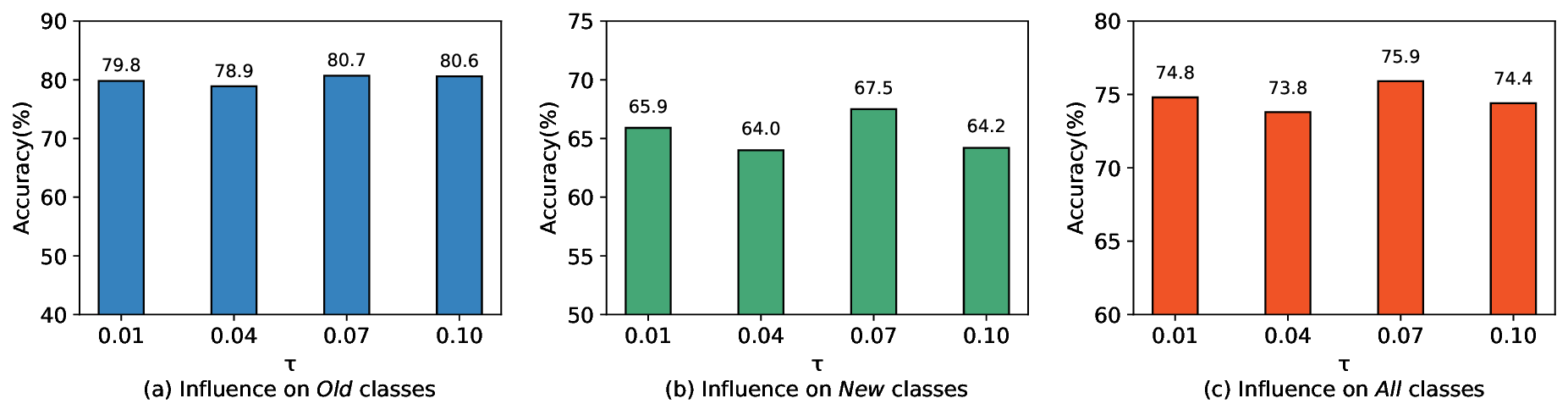}
    \caption{Clustering accuracy in the last incremental session under different temperature ${\tau}$.}
    \label{fig:NCENet_temperature}
\end{figure*}

\begin{figure*}
    \centering
    \includegraphics[width=1.9\columnwidth]{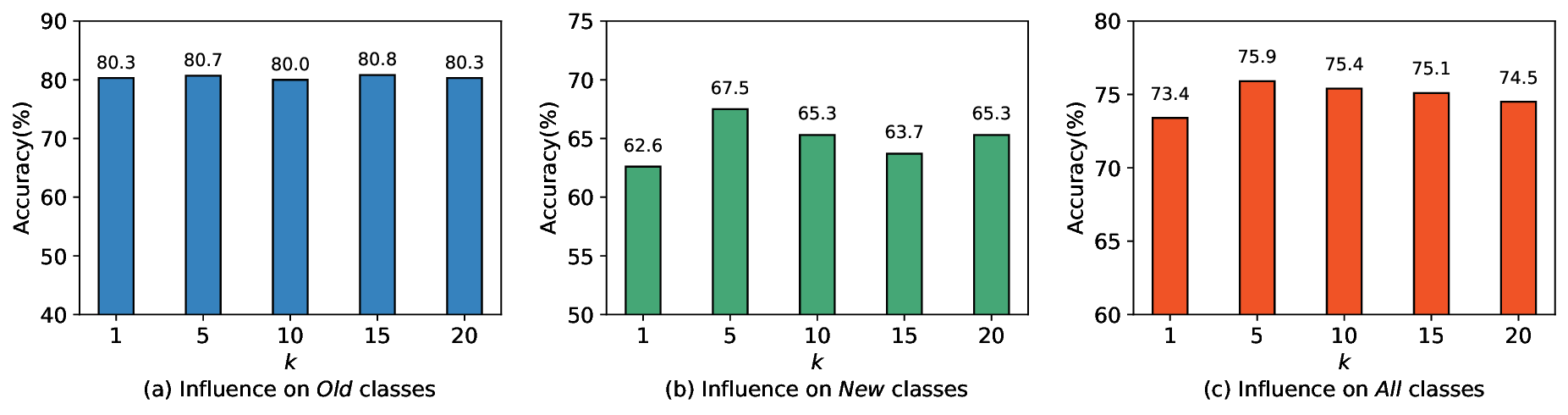}
    \caption{Clustering accuracy in the last incremental learning session under different numbers of selected neighbors. A relatively larger number can help our proposed method achieve better performance.}
    \label{fig:NCENet_neighbors}
\end{figure*}

\subsection{Ablation study}

Our proposed NCENet relies on Neighborhood Commonality-aware Representation Learning (NCRL) to enhance the novel class discovery ability and Bi-level Contrastive Knowledge Distillation (BCKD) to mitigate the catastrophic forgetting problem. To validate the effectiveness of each module, we conduct several ablation studies on CIFAR10 and report corresponding clustering accuracy in Table~\ref{tab:ablat_NCENet}. From the table, we can see that compared to the performance given by using both NCRL and BCKD (row 3), though removing NCRL (row 1) leads to a performance improvement on old class clustering accuracy (\textit{Old}), it drops the new class clustering accuracy (\textit{New}) by a relatively larger margin in each session, which results in performance degradation on all class clustering accuracy (\textit{All}). Particularly, the $m$N/$m$A without using NCRL is 78.77\%/94.57\% while that given by using NCRL is 89.36\%/92.56\%, this indicates that NCRL is pivotal in novel category discovery. Further, though removing BCKD (row 2) improves the clustering accuracy on old classes, it drops the clustering accuracy on new classes and all classes in each session. Especially, removing BCKD drops the $m$O from 95.87\% to 90.24\% and $m$A from 94.57\% to 91.93\%, this suggests that BCKD plays a key role in old knowledge retention. In summary, experimental results shown in Table~\ref{tab:ablat_NCENet} show that our proposed NCRL and BCKD are both effective. Further, combining NCRL and BCKD can achieve better entire clustering accuracy than using one of them solely.

\section{Discussion}\label{sec:exp}

\subsection{Discussion about hyperparameter ${\lambda}_b$}

To investigate the influence of ${\lambda}_b$ used to balance contributions of NCRL and BCKD, we vary the value of ${\lambda}_b$ across $\{0.1, 0.3, 0.5, 0.7, 0.9\}$ and report the corresponding clustering accuracy on \textit{Old}/\textit{New}/\textit{All} classes in Figure~\ref{fig:NCENet_plast}. As depicted in Figure~\ref{fig:NCENet_plast}(a), employing a smaller ${\lambda}_b$ is more beneficial for preserving the model’s clustering accuracy on \textit{Old} classes. In contrast, as shown in Figure~\ref{fig:NCENet_plast}(b), a larger ${\lambda}_b$ yields superior clustering accuracy on \textit{New} classes during the first two incremental sessions. However, the clustering performance discrepancy between different ${\lambda}_b$ diminishes as the learning progresses. The main reason we guess is that though the improvement on \textit{New} classes given by using a smaller ${\lambda}_b$ value is smaller than that given by using a larger smaller ${\lambda}_b$ value, using a smaller ${\lambda}_b$ value can achieve better old knowledge retention, which contributes to unleashing the potential of NCRL for learning new knowledge. Consequently, as shown in Figure~\ref{fig:NCENet_plast}(c), using a relatively smaller ${\lambda}_b$ value helps our proposed NCENet achieve better clustering accuracy on \textit{All} classes. Particularly, setting the value of ${\lambda}_b$ to 0.1 is an optimal choice for our proposed method.

\subsection{Discussion about NCRL}
\label{dis_col}

\noindent \textbf{Temperature $\tau$} 
In NCRL, we use a temperature $\tau$ to sharpen the prediction distribution of one of the augmentation views. To explore the influence of $\tau$, we change $\tau$ among $\{0.01, 0.04, 0.07, 0.10\}$ and report corresponding clustering accuracy on \textit{Old}/\textit{New}/\textit{All} classes of last session in Figure~\ref{fig:NCENet_temperature}. From Figure~\ref{fig:NCENet_temperature}(a), we can see that increasing $\tau$ from 0.01 to 0.04 results in a clustering accuracy degradation on \textit{Old} classes. Conversely, increasing $\tau$ from 0.04 to a larger value boosts the clustering accuracy on \textit{Old} classes by a relatively larger margin. However, as shown in Figure~\ref{fig:NCENet_temperature}(b), though setting $\tau$ to 0.07 and 0.1 both achieves a satisfactory clustering accuracy on \textit{Old} classes, setting $\tau$ to 0.07 achieves better clustering accuracy on \textit{New} classes. Overall, as depicted in Figure~\ref{fig:NCENet_temperature}(c), setting $\tau$ to 0.07 helps our proposed method achieve the best clustering performance.

\noindent \textbf{The number of neighbors in NCRL.} 
To explore the influence of different numbers of neighbors on the model's clustering performance, we set $k$ to different values and report the corresponding clustering accuracy on \textit{Old}/\textit{New}/\textit{All} classes of last session in Figure~\ref{fig:NCENet_neighbors}. As we can see from Figure~\ref{fig:NCENet_neighbors}(a), the clustering accuracy on \textit{Old} classes is relatively stable across different $k$ values. However, as shown in Figure~\ref{fig:NCENet_neighbors}(b), setting $k$ to a relatively larger value achieves better clustering accuracy on \textit{New} classes. We speculate that using a single instance may inadequately represent the local commonality, thereby compromising the effectiveness of NCRL. Further, compared to other larger $k$ values, changing $k$ from 1 to 5 achieves the most significant clustering accuracy improvement, approximately 5$\%$. The main reason we guess is that a large $k$ value may introduce noise to commonality representation, which also undermines the effectiveness of NCRL. Overall, as shown in Figure~\ref{fig:NCENet_neighbors}(c), setting $k$ to 5 helps our proposed method achieve the best clustering accuracy on \textit{All} classes.

\begin{figure*}
    \centering
    \includegraphics[width=1.9\columnwidth]{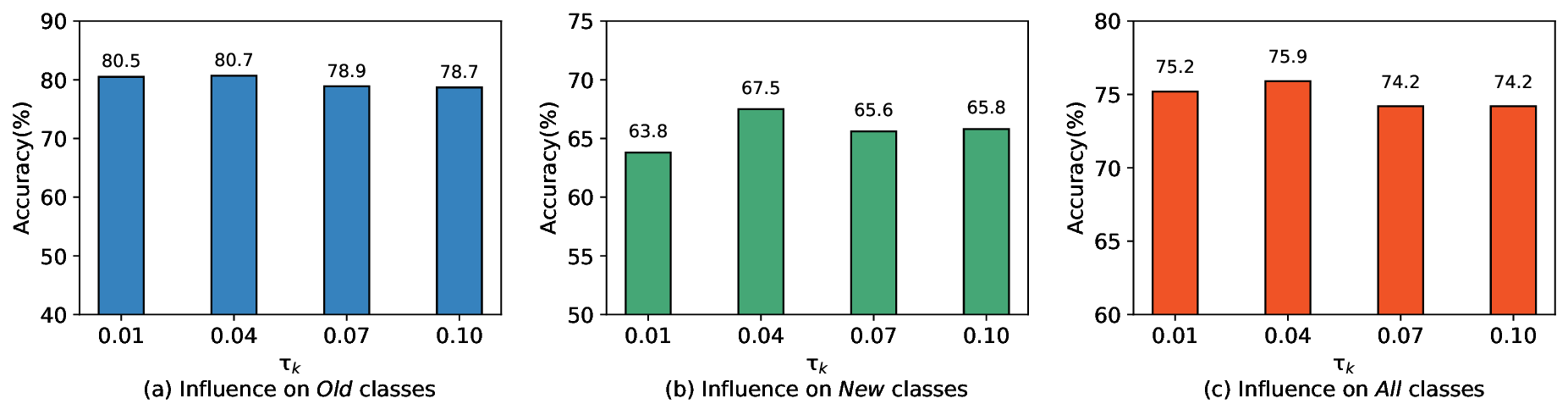}
    \caption{Clustering accuracy in the last incremental learning session under different ${\tau}_k$ used in our proposed BCKD. Using a relatively smaller temperature helps improve the model's clustering performance.}
    \label{fig:NCENet_temperature}
\end{figure*}

\begin{table}[ht]
    \centering
    \caption{Clustering accuracy of last session under different neighbor selection strategies, where $\alpha$ refers to the threshold used to select neighbors.}
    \label{tab:NCENet_neighbors_select}
    
    \setlength{\tabcolsep}{2.2mm}{
    \begin{tabular}{l|cc|ccc}
    \toprule
    Exp. & Strategy & Hyper. & Old & New & All \\
    \midrule
    1) & \multirow{4}{*}{Threshold} & {$\alpha =0.6$} & {79.59} & {64.45} & {73.80} \\
    2) & {} & {$\alpha =0.7$} & \textbf{80.96} & {63.25} & {74.54} \\
    3) & {} & {$\alpha =0.8$} & {80.54} & {63.10} & {74.79} \\
    4) & {} & {$\alpha =0.9$} & {79.51} & \textbf{67.50} & {74.42} \\
    \midrule
    5) & Number & $k=5$ & 80.69 & 67.45 & \textbf{75.88}  \\
    \bottomrule
         
    \end{tabular}}
\end{table}

\noindent \textbf{Neighbor selection strategies.} 
To explore whether using a threshold is more optimal than using a fixed number to select neighbors, we use the performance given by using a fixed number 5 to select neighbors as the baseline, and then switch the neighbor selection strategy to a threshold-based strategy. To make a comprehensive and convincing comparison, we vary the value of threshold $\alpha$ across $\{0.6, 0.7, 0.8, 0.9\}$ and report the corresponding clustering accuracy on  \textit{Old}, \textit{New} and \textit{All} classes. As we can see from Table~\ref{tab:NCENet_neighbors_select}, compared to the baseline,  though setting $\alpha$ to  0.7 achieves better performance on \textit{Old} classes, it drops the clustering performance on \textit{New} classes by a relatively larger margin. Furthermore, compared to the baseline, though setting setting $\alpha$ to 0.9 achieves competitive performance. However, it results in a poorer clustering performance on \textit{Old} classes. In summary, using a fixed number is more helpful than using a threshold to select neighbors for our proposed method.

\begin{table}[ht]
    \centering
    \caption{Averaged clustering accuracy under different initialization methods, where $mT$ is the mean time cost in each training epoch. }
    \label{tab:initialization}
    
    \setlength{\tabcolsep}{1.2mm}{
    \begin{tabular}{l|l|ccc|c}
    \toprule
    Exp. & Initialization & mO & mN & mA & mT\\
    \midrule
    1) & random                                                 & 81.44 & 65.12 & 77.17 & 0 \\
    2) & KMeans                                                 & 81.54 & \textbf{70.49} & 78.14 & 43.06s \\
    \midrule
    3) & Commonality (Ours)                                      & \textbf{81.62} & {70.01} & \textbf{78.28} & 0.11s  \\
    \bottomrule
    \end{tabular}}
\end{table}

\noindent \textbf{Analysis of initialization.}
Different from previous methods that use randomly initialized classification heads to generate prediction distributions, NCRL exploits local commonalities to produce prediction distributions. To validate the effectiveness of this approach, we use the results obtained from generating prediction distributions with randomly initialized classification heads as the baseline (Random) and then switch to other methods. As shown in Table \ref{tab:initialization}, compared to the baseline, using KMeans for clustering and employing the centers of each cluster to generate prediction distributions can achieve better clustering accuracy on \textit{New} classes but comes with an additional time cost of 43.06s. Compared to the baseline, using local commonalities to generate prediction distributions also improves the clustering accuracy on \textit{New} classes. Although using local commonalities performs relatively worse on new classes compared to the results obtained with KMeans, it achieves a certain degree of improvement in the clustering accuracy on \textit{Old} and \textit{All} classes. Meanwhile, it only adds 0.11s to the time consumption. It is worth noting that KMeans often requires a predetermined number of categories.

\begin{figure}[ht]
    \centering
    \includegraphics[width=.99\columnwidth]{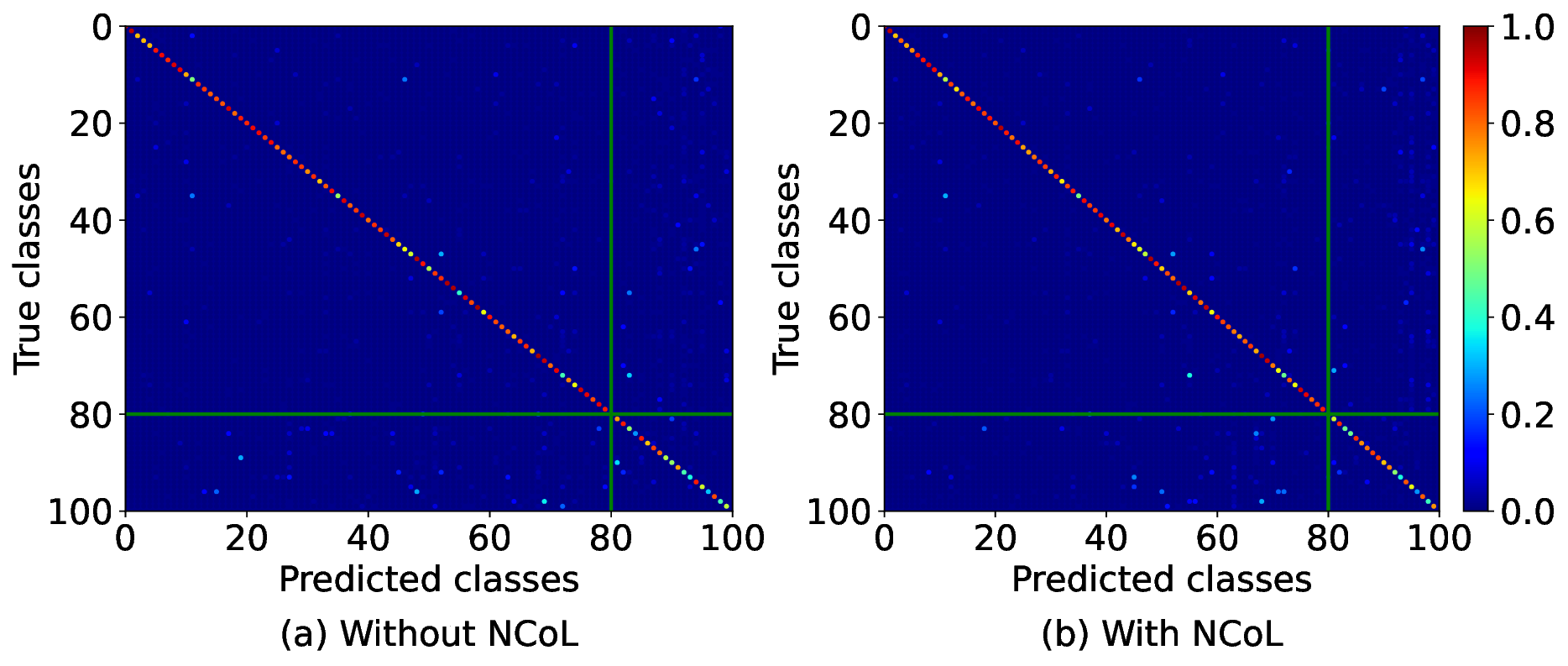}
    \caption{Clustering accuracy without and with using NCRL module. }
    \label{fig:NCENet_cm}
\end{figure}

\noindent{\bf Visualization.} To further give an insight into our proposed NCRL, Figure \ref{fig:NCENet_cm} shows the confusion matrices obtained by removing NCRL and using NCRL, we can see that both removing and using NCRL can achieve satisfactory label assignment results on the old classes. However, on the new classes, compared to the visualization results obtained by removing NCRL, the visualization results using NCRL have more red dots on the diagonal, indicating that more instances within each category are correctly assigned labels. Overall, the visualization results of this experiment further demonstrate that NCRL can effectively improve the model's performance on new classes.

\subsection{Discussion about BCKD}
\label{dis_bckd}

\noindent \textbf{Temperature ${\tau}_k$ in BCKD.}
To investigate the influence of temperature ${\tau}_k$ used in our proposed Bi-level Contrastive Knowledge Distillation (BCKD), we change the value of ${\tau}_k$ among $\{0.01, 0.04, 0.07, 0.10\}$ and report the clustering accuracy on \textit{Old}/\textit{New}/\textit{All} classes given by different ${\tau}_k$ values in Figure~\ref{fig:NCENet_temperature}. As shown in Figure~\ref{fig:NCENet_temperature}(a), we find that setting a relatively smaller ${\tau}_k$ value helps our proposed NCENet achieves better clustering accuracy on \textit{Old} classes. Conversely, as we can see from Figure~\ref{fig:NCENet_temperature}(b), setting a relatively larger ${\tau}_k$ value helps our proposed NCENet achieves better clustering accuracy on \textit{New} classes. Overall, as shown in Figure~\ref{fig:NCENet_temperature}(c), setting the value of  ${\tau}_k$ to 0.04 achieves the best clustering accuracy on \textit{All} classes.

\begin{table}[ht]
    \centering
    \caption{Clustering accuracy in the last incremental learning session under different knowledge distillation losses. }
    \label{tab:NCENet_losses}
    \setlength{\tabcolsep}{3mm}{
    \begin{tabular}{l|l|ccc}
    \toprule
    Exp. & Distillation loss & Old & New & All\\
    \midrule
    1) & MSE                                                 & 79.43 & 61.40 & 72.99 \\
    2) & KL divergence                                       & 78.06 & 66.35 & 73.50  \\
    \midrule
    3) & BCKD-$\mathcal{L}_{\text{sa}}$                           & 79.44 & 67.25 & 73.98  \\
    4) & BCKD-$\mathcal{L}_{\text{ta}}$                           & \textbf{81.05} & 63.05 & 75.09  \\
    5) & BCKD-$\mathcal{L}_{\text{sa}}$+$\mathcal{L}_{\text{ta}}$ & {80.69} & \textbf{67.45} & \textbf{75.88}  \\
    \bottomrule
    \end{tabular}}
\end{table}

\noindent \textbf{Different knowledge distillation losses.} To further validate the effectiveness of our proposed Bi-level Contrastive Knowledge Distillation (BCKD), we compare BCKD with the commonly used knowledge distillation losses, including MSE and KL divergence. As shown in Table~\ref{tab:NCENet_losses}, using MSE to perform knowledge distillation achieves a better clustering accuracy on \textit{Old} classes than KL divergence, but the clustering accuracy on \textit{New} classes is dropped by a relatively larger margin. Using our proposed $\mathcal{L}_{\text{sa}}$/$\mathcal{L}_{\text{ta}}$ achieves better clustering accuracy on \textit{Old} classes than MSE, this demonstrates that using contrastive learning to perform knowledge distillation is more helpful for our proposed method to achieve better old knowledge retention. Meanwhile, we observe that using $\mathcal{L}_{\text{ta}}$ results in a better clustering accuracy on \textit{Old} classes than using $\mathcal{L}_{\text{sa}}$, but it drops the clustering accuracy on \textit{New} classes by a relatively larger margin. The main reason may be that only perceiving the learned knowledge leads to an over-constrained issue which undermines the new knowledge learning ability. Ultimately, compared to the clustering performance given by using only $\mathcal{L}_{\text{ta}}$, though combing $\mathcal{L}_{\text{sa}}$ and $\mathcal{L}_{\text{ta}}$ drops the clustering accuracy on \textit{Old} classes slightly, it improves the clustering accuracy on \textit{New} classes by a relatively larger margin and achieves the best clustering accuracy on \textit{All} classes, this implies that introducing the learning knowledge can mitigate the over-constrained issue.

%% file: sections/6.conclusion.tex
\section{Conclusion}\label{sec:conclusion and future work}

In this paper, we solve the challenging C-GCD problem from the perspective of representation learning and propose a Neighborhood Commonality-aware Evolution Network. Firstly, we devise a Neighborhood Commonality-aware Representation Learning module that incorporates local commonalities obtained from different neighborhoods with the self-distillation technique to learn discriminative representations for novel classes. Secondly, we devise a Bi-level Contrastive Knowledge Distillation module that exploits student-anchored contrastive knowledge distillation and teacher-anchored contrastive knowledge to maintain the model's representation ability for old classes. Extensive experimental results demonstrate the state-of-the-art performance of our proposed method on multiple benchmark datasets.

\noindent{\bf Limitation and Future Work.} Common incremental settings are more than a few incremental steps. However, this work only deals with the incremental setting with a maximum of 5 incremental steps, which are relatively short, thus limiting the applications. How to model C-GCD with long incremental steps remains an interesting problem.